\pdfoutput=1

\documentclass[11pt]{article}

\usepackage[preprint]{acl}

\usepackage{times}
\usepackage{latexsym}
\usepackage{booktabs}
\usepackage{multirow}
\usepackage{float}
\usepackage{amsmath}
\usepackage[T1]{fontenc}
\usepackage{float}

\usepackage[utf8]{inputenc}

\usepackage{graphicx}
\usepackage{microtype}

\usepackage{inconsolata}

\title{LANS: A Layout-Aware Neural Solver for Plane Geometry Problem}

\author{
    Zhong-Zhi Li\textsuperscript{1,2}\thanks{~~~Equal Contribution} , Ming-Liang Zhang\textsuperscript{1,2}\footnotemark[1], Fei Yin\textsuperscript{1,2}, Cheng-Lin Liu\textsuperscript{1,2}\thanks{~~~Corresponding Author}\\
    School of Artifcial Intelligence, University of Chinese Academy of Sciences\textsuperscript{1} \\
    MAIS, Institute of Automation of Chinese Academy of Sciences\textsuperscript{2}, \\
    \{lizhongzhi2022, zhangmingliang2018\}@ia.ac.cn, \\
    \{fyin, liucl\}@nlpr.ia.ac.cn
}

\begin{document}
\maketitle
\begin{abstract}
Geometry problem solving (GPS) is a challenging mathematical reasoning task requiring multi-modal understanding, fusion, and reasoning. Existing neural solvers take GPS as a vision-language task but are short in the representation of geometry diagrams that carry rich and complex layout information. In this paper, we propose a layout-aware neural solver named LANS, integrated with two new modules: multimodal layout-aware pre-trained language module (MLA-PLM) and layout-aware fusion attention (LA-FA). MLA-PLM adopts structural-semantic pre-training (SSP) to implement global relationship modeling, and point-match pre-training (PMP) to achieve alignment between visual points and textual points. LA-FA employs a layout-aware attention mask to realize point-guided cross-modal fusion for further boosting layout awareness of LANS. Extensive experiments on datasets Geometry3K and PGPS9K validate the effectiveness of the layout-aware modules and superior problem-solving performance of our LANS solver, over existing symbolic and neural solvers. The code will be made public available soon.
\end{abstract}

\section{Introduction}
Automatic geometry problem solving (GPS) is a long-standing and challenging research topic in both computer vision and natural language processing communities \cite{Bobrow1964,Chou1996,Seo2015}.  Each geometry problem consists of a geometry diagram and a textual problem in different modal forms, complementing each other.  GPS necessitates comprehensive mathematical reasoning and multi-modal understanding, making it a pivotal testbed for evaluating the high-level multimodal reasoning ability of artificial intelligence. Past research works of GPS were mainly focused on \textit{symbolic solvers} \cite{Seo2015, Sachan2017, Lu2021}, which are criticized in respect of complex rules and poor adaptability. With the development of deep learning, \textit{neural solvers} \cite{Chen2021, Chen2022, Zhang2023, GAPS}, treating GPS as a special vision-language reasoning task, have attracted dominant attention recently. 
\begin{figure}[t]
    \begin{center}
    \includegraphics[width=0.95\columnwidth]{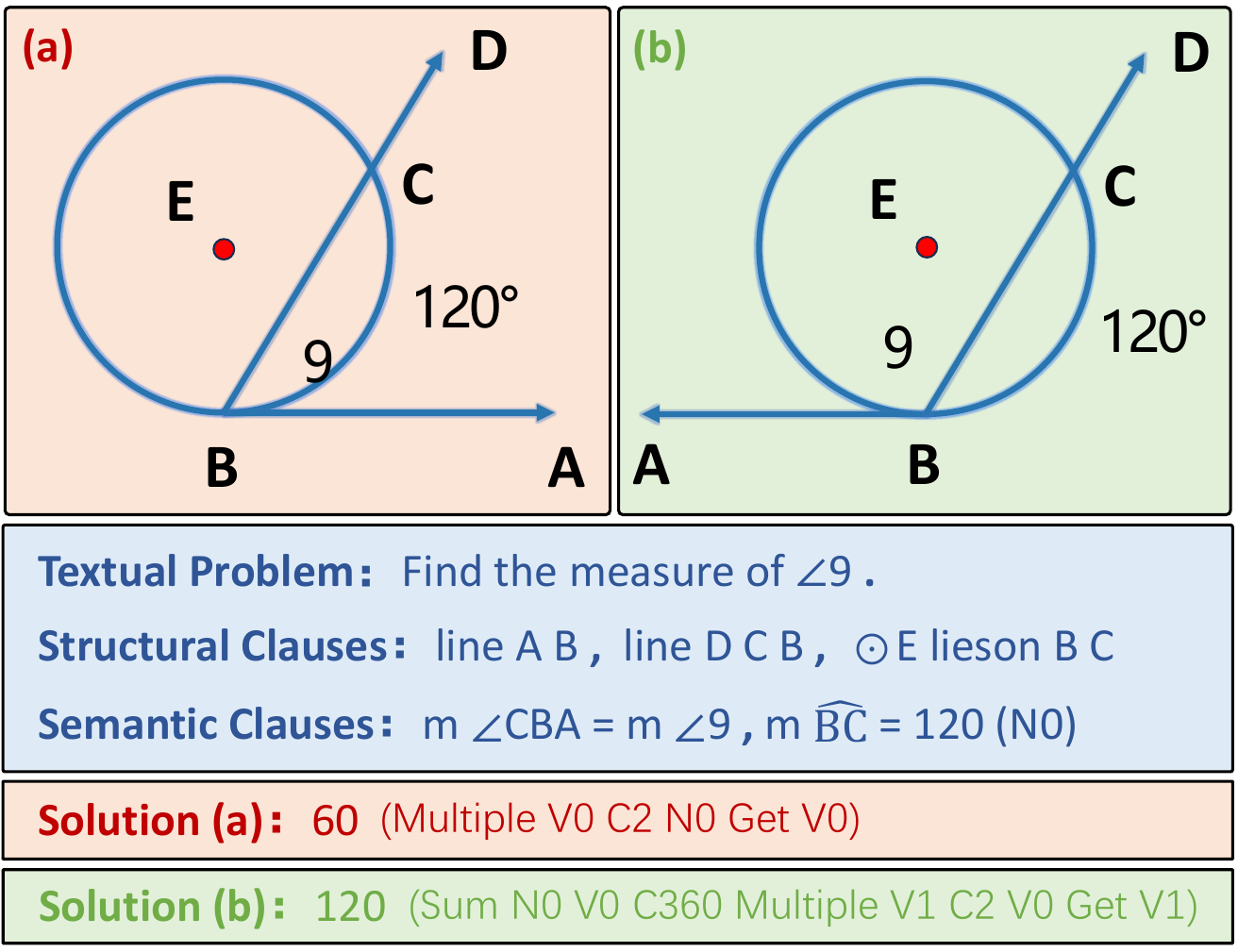} 
    \end{center}
    \vspace{-0.2cm}
    \caption{Examples of plane geometry problems. The geometry diagrams (a) and (b) share the same textual problem, structural clauses, and semantic clauses but have different solutions, where structural clauses and semantic clauses are parsed from diagrams. Layout information plays a crucial role in this situation.}
    \vspace{-0.2cm}
    \label{fig:example}
\end{figure}

Layout information is typically defined as positional coordinates of elements such as text, paragraphs, tables, and figures within images \cite{Xu2020, gupta2021layouttransformer}. Supplying layout details for elements in document images facilitates parsing reading sequences, executing information extraction, and enhancing document comprehension \cite{DocFormer, wang-layoutreader-21, bros-2022-aaai}. In the layout of geometric diagram, the coordinate positions of geometric points and symbols play a crucial role in understanding the elements within geometric diagrams. For example, the coordinate positions of geometric symbols "A" as shown in Figure \ref{fig:example}, determine which geometric points are named A, while the coordinate position of the non-geometric symbol "$120^\circ$" determines the numerical assignment of $\angle ABD$ instead of other angle.

Despite considerable efforts devoted to constructing proficiently crafted representations for geometric diagrams, the explicit fusion of positional information into geometric diagrams remains unexplored. Existing neural solvers have adopted different diagram representation schemes, such as \textit{feature maps} \cite{Chen2021, Cao2022, Ning2023}, \textit{image patches} \cite{Chen2022, Ning2023} and \textit{textual clauses} \cite{Lu2021, Zhang2023}. For methods based on \textit{image patches} and \textit{feature maps}, several representative geometric problem solvers have employed extensive pre-training strategies, such as jigsaw location prediction \cite{Chen2021}, geometry elements prediction \cite{Chen2021}, masked image modeling \cite{Ning2023}, and character alignment 
 \cite{Ning2023}, to bridge the gap between geometric and natural scene images \cite{Peter2018, Zhou2019, Ding2022}. Although rough image pre-training methods have achieved some effectiveness, they often fail to capture finer-grained details. Conversely, methods based on \textit{text clauses} extract the crucial structural and semantic information of geometric problems in the form of clauses. Currently, clause-based approaches yield superior inference results through clause-based deductive reasoning \cite{Lu2021} or clause pre-training \cite{Zhang2023}. We attribute this to the structured nature of clauses, which makes them more adept at capturing structural information in geometric problems. For example, The structural clause ``line B C D" describes a structural relationship that points ``B", ``C" and ``D" lie on one line in order. The semantic clause ``m $\widehat{\rm{BC}}$ = 120" illustrates a semantic relationship for the degree of arc ``$\widehat{\rm{BC}}$" and text ``120$^{\circ}$".   

Although the textual clauses are capable of capturing the primary layout relationships within the images, they lose significant spatial information during the conversion process of diagram parsing \cite{Lu2023,trinh2024solving}. They cannot distinguish the geometry diagram (a) and (b) displayed in Figure 1 because of loss of position information. For example, “$\angle$CBA" in Figure \ref{fig:example}(a) and (b) need the spatial relationship to determine whether it is acute or obtuse. The lack of positional indicators for geometry elements (such as "A," "B," etc.) makes it challenging for neural solvers based on text clauses to distinguish between these ambiguous scenarios.

\begin{figure}[t]
    \begin{center}
    \includegraphics[width=0.95\columnwidth]{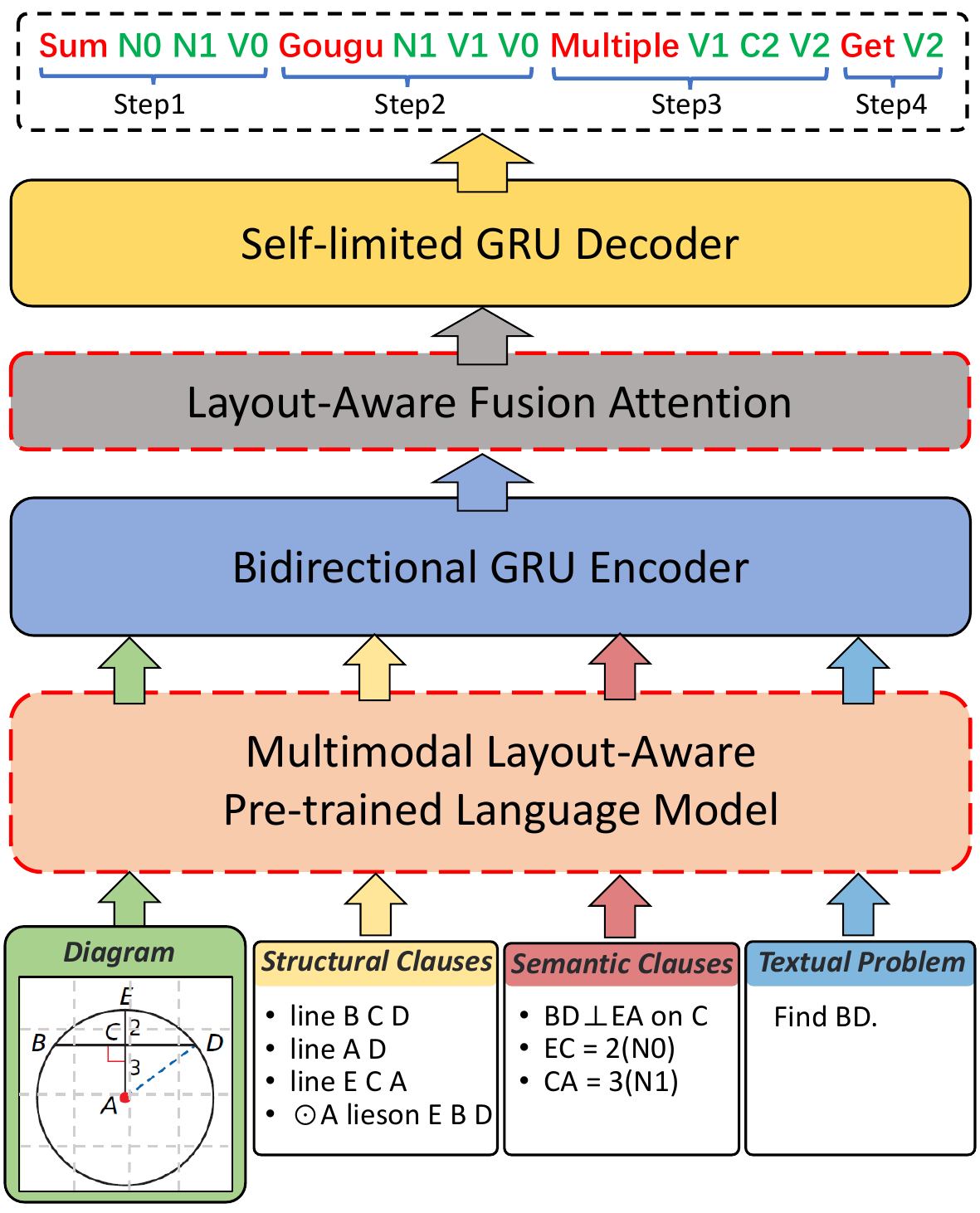} 
    \end{center}
    \vspace{-0.3cm}
    \caption{Overview of LANS model. The red dotted boxes are our newly proposed modules in comparison to PGPSNet \cite{Zhang2023}.}
    \label{fig:network}
\end{figure}

Considering the under-representation of geometry diagrams, we propose a layout-aware neural solver called LANS. LANS inputs the diagram image, the textual clauses parsed from a diagram, and the textual problem, and outputs the explainable solution program to solve the geometry problem. 
As shown in Figure \ref{fig:network}, two new modules, multimodal layout-aware pre-trained language model (MLA-PLM) and layout-aware fusion attention (LA-FA), are proposed to endow LANS with layout awareness. We introduce a point-match pre-training (PMP) method within MLA-PLM. This method, based on contrastive learning, aims to model the relationship between text clauses and diagrams using layout information in a data-efficient  manner. When integrated with structural-semantic pre-training (SSP) in PGPS solver \cite{Zhang2023}, it shows promising outcomes. Then, to better utilize pre-trained multimodal representations, LA-FA module with the layout-aware attention mask is employed in LANS to fuse the diagram and text clauses representation via point positions. LA-FA further enhances the layout awareness in cross-modal fusion. 

The contributions of this work are summarized in four folds: (1) We propose a layout-aware neural solver LANS for GPS, which can represent and fuse geometry diagrams effectively. (2) We introduce the MLA-PLM module with two pre-training strategies SSP and PMP, realizing global relationship modeling and cross-modal alignment of point primitives. (3) We design the LA-FA module, equipped with a layout-aware attention mask directed by point positions, to further strengthen the layout awareness of LANS. (4) Our LANS outperforms existing symbolic solvers, neural solvers, and current multimodal large models significantly on Geometry3K and PGPS9K datasets.

\section{Related Work}

\subsection{Geometry Problem Solving} 
GPS is a special type of multimodal reasoning that examines geometric spatial structure cognition and mathematical logical reasoning, and also requires the application of geometric theorem knowledge, which make it highly challenging.
Existing works of GPS can be classified into two categories: symbolic solvers and neural solvers. The symbolic solvers \cite{Seo2015, Sachan2017, Lu2021, Peng2023} parse the diagram and textual problem into a unified formal language first, and then perform symbolic reasoning by path search and condition matching based on the geometric theorem knowledge. However, symbolic solvers are carefully designed with complex rules and are hard to extend. The neural solvers treat GPS as a visual question answering task and design a special interpretable program to represent the problem-solving process. NGS \cite{Chen2021} and Geoformer \cite{Chen2022} use auxiliary self-supervised tasks such as location prediction, elements prediction, and knowledge classification to boost cross-modal semantic representation. PGPSNet \cite{Zhang2023} expresses the geometry diagram with textual clauses and fuses multi-modal information through structural and semantic pre-training, data augmentation, and self-limited decoding. SCA-GPS \cite{Ning2023} tries to align characters in text and diagram and enhance the diagram understanding through multi-label classification and masked image modeling pre-training. Although existing neural solvers have achieved impressive performance, they are still coarse-grained at the modal understanding and fusion, especially for geometry diagrams with complex layouts. In this paper, we propose a layout-aware neural solver to improve the understanding and fusion of geometry diagrams and therefore promote GPS.
\subsection{Multimodal Pre-training \& Layout-Aware Learning}
Multimodal pre-training realizes alignment and understanding between different modalities by a series of designed auxiliary tasks and then applies to the specific downstream tasks. Common strategies involve image-text contrastive learning \cite{radford2021learning}, image-text matching \cite{Kim2021}, image-grounded text generation \cite{cho2021vlt5}, and masked object classification \cite{Li2020}. With a large amount of pre-training data, these strategies exhibit good performance in multimodal tasks for natural images. However, their alignment methods are coarse-grained and straightforward and do not fit for complex multi-level and fine-grained tasks. Most relevant to our work is the research on document analysis \cite{Liu2023FrontiersOI}. Existing advanced document pre-training methods \cite{Xu2020, Xu2021} incorporate textual and visual blocks with fine-grained position embeddings, and adopt masked visual-language modeling and text-image alignment to pretrain document layout, whereas they still do not apply to GPS due to the specificity of geometry objects and small-scale of GPS datasets. 
DocFormer \cite{DocFormer} and LayoutReader \cite{wang-layoutreader-21} employ meticulously designed attention mechanisms targeting information within text boxes to enhance their perception abilities regarding document content. Our LANS proposes targeted and data-efficient pre-training methods and a geometry layout-aware attention to implement geometry layout awareness. 

\section{Method}
Before presenting the neural solver model, we first describe the formal definition of GPS task here. Given a geometry problem $P$ including a geometry diagram $D$ and a textual problem $T_{prob}$, the goal is to solve the problem by applying geometric knowledge and obtaining the solution steps $S$, formulated as $P=\left\{D, T_{prob}\right\} \Rightarrow S$. Then solution steps are verified in the form of fill-in-the-blank, multiple-choice, or logical reasons.

\begin{figure*}[t]
    \begin{center}
    \includegraphics[width=1.9\columnwidth]{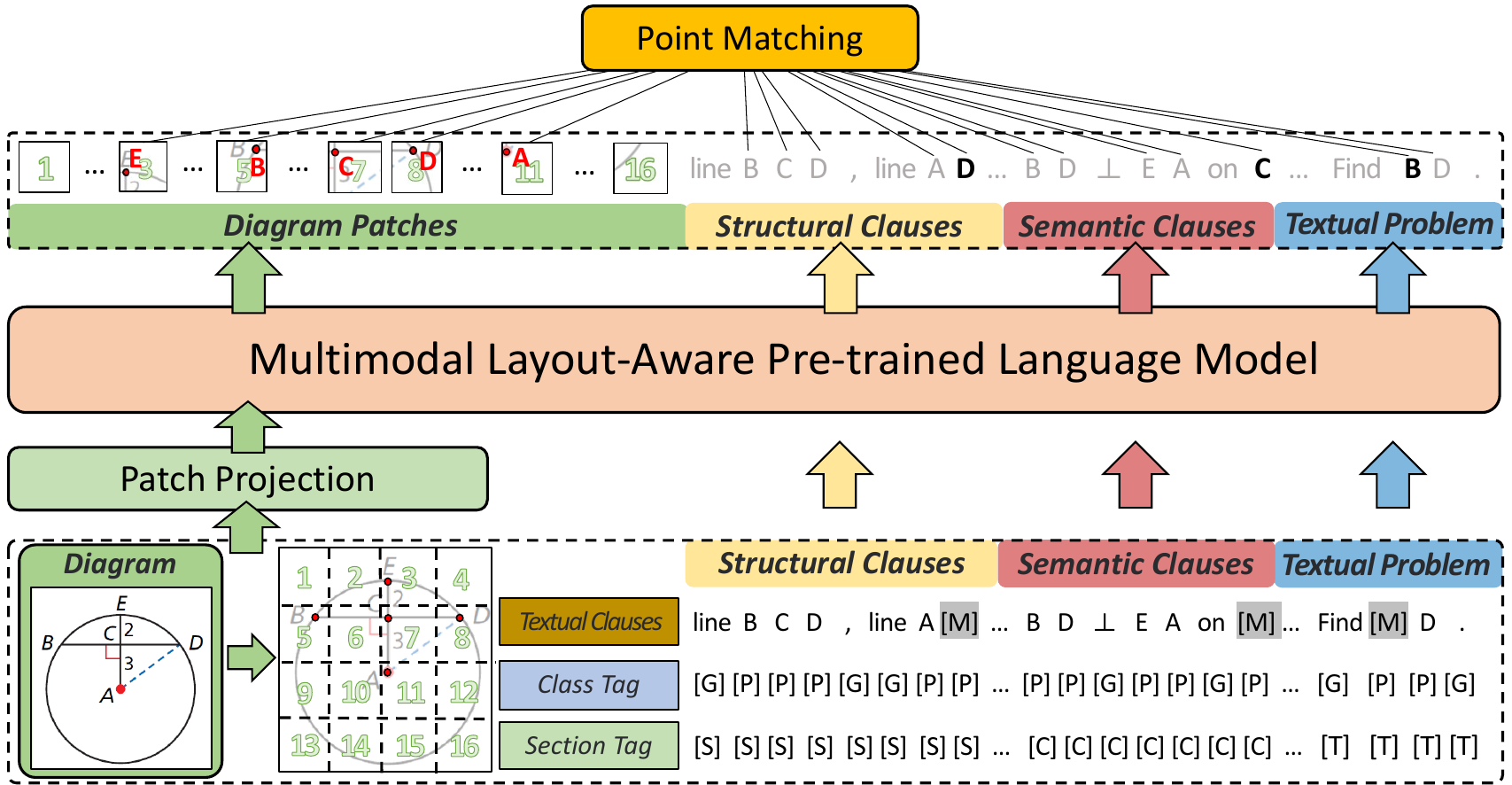} 
    \end{center}
    \caption{Pipeline of multimodal layout-aware pre-training. The geometry problem is the same as that in Figure \ref{fig:network}. [M] denotes mask tokens. Class tags and section tags are the same as \cite{Zhang2023}.}
    \label{fig:pretraining}
\end{figure*}

\subsection{Overall Framework}
To fully understand and represent the geometry diagram, we propose a layout-aware neural solver called LANS as displayed in Figure \ref{fig:network}. First, the diagram is parsed into the textual clauses using the geometry diagram parser PGDPNet \cite{Zhang2022}, where the structural clauses $T_{stru}$ describe the connection relations among geometric primitives and the semantic clauses $T_{sem}$ depict the semantic relations between non-geometric primitives and geometric primitives \cite{Zhang2023}. Besides, the visual information of diagram image is represented as patches. Therefore, the input of LANS could be further expressed as $\left\{D\!=\!\left\{d_i\right\}_{i=1}^{N_D}, T\!\!=\!\!\left\{T_{stru}, T_{sem}, T_{prob}\right\}\!\!=\!\!\left\{t_j\right\}_{j=1}^{N_T}\right\}$ after token concatenation, where $N_D$ is the diagram patch number and $N_T$ is the text token number. Then, these modal tokens are fed into the multimodal layout-aware pre-trained language model (MLA-PLM) and input into the bidirectional GRU encoder to perform fusion encoding. Next, the mixed encoding context $H=\left\{h_i\right\}_{i=1}^{N_D\!+\!N_T}$ leverages the layout-aware fusion attention (LA-FA) to further boost diagram layout awareness. Finally, the enhanced context is decoded by the self-limited GRU decoder and generates the sequential solution program $S$ in the manner of autoregressive.


\subsection{Multimodal Layout-Aware Pre-training}
Geometry problems are often solved by humans by depicting the geometric structure in visual form no matter whether it has the geometry diagram or not. Previous neural geometric solvers, such as the NGS \cite{Chen2021}, PGPSNet \cite{Zhang2023} and SCA-GPS \cite{Ning2023}, do not utilize the diagram layout adequately, thus resulting in unsatisfactory performance of GPS. In this paper, we propose the multimodal layout-aware pre-trained language model (MLA-PLM), with two pre-training strategies: structural-semantic pre-training (SSP) and point matching pre-training (PMP) illustrated in Figure \ref{fig:pretraining}, to boost the diagram layout-aware ability during the pre-training stage.
\paragraph{Revisit Structural-Semantic Pre-training} To enable the multimodal pre-training module to comprehend text clauses and gain a preliminary understanding of the content and layout of geometry diagrams, we adopted the structural-semantic pre-training (SSP) \cite{Zhang2023} method used in PGPS. MLA-PLM is trained to recover the masked text in a unified text generation manner, and the training loss denotes as $L_{S\!S\!P}$. Concretely, inputs of MLA-PLM include the diagram patch embeddings $e_i^{D}$ and textual token embeddings $e_j^{T}$, where $e_i^{D}$ is obtained via patch projection and patch-level positional encoding, and $e_j^{T}$ fuses not only positional encoding but also embedding of class tag and section tag following \cite{Zhang2023} as: 
\begin{equation}
    \begin{aligned}
    e_i^{D}\! &=\! \text{PatchProj}(d_i) \!+\! \text{PosEmb}(i), \;  1 \!\leq\! i \!\leq N_D \\
    e_j^{T} \!&=\! \text{TokenEmb}(t_j) \!+\! \text{PosEmb}(j) \!+\! \text{ClassEmb}(t_j) \\ 
    & \qquad \qquad \qquad \quad + \text{SectEmb}(t_j), \;1 \leq j \leq N_T
    \end{aligned},
\end{equation}
where $\text{PosEmb}(*)$ is the sequential position encoding of sequences instead of the spatial position of the diagram layout. The concatenated $e_i^{D}$ and $e_j^{T}$ are modeled by MLA-PLM and then output ${e'}_i^{D}$ and ${e'}_j^{T}$. For SSP in MLA-PLM, we mask 30\% of text tokens $t_j$ with mask token $[M]$ following \cite{cho2021vlt5} but keep tags unchanged. 
\paragraph{Point-Match Pre-training} We propose the PMP based on contrastive, learning modeling to achieve cross-modal alignment between visual points (one type of geometric primitives in the diagram) and textual tokens of the points. For PMP, we match image patches and points inside image patches with the cosine contrastive loss \cite{he2020momentum,grill2020bootstrap} as follows:
\begin{equation}
  L_{P\!M\!P}=\frac{-1}{|\mathcal{P}|}\sum_{j \in \mathcal{P}}\log\frac{\exp(\cos\langle{e'}_j^{T}, {e'}_+^{D}\rangle/\tau)}{\sum_{i=1}^{N_D}\exp(\cos\langle{e'}_j^{T}, {e'}_i^{D}\rangle/\tau)},
\end{equation}
where $\mathcal{P}=\{j \;|\; \text{Class}(t_j) = \text{[P]}, \; 1 \leq j \leq N_T\}$ is the index list of text tokens corresponding to points, ${e'}_+^{D}$ is the embedding of the diagram patch that the point $t_j$ is located in, and $\tau$ is the temperature coefficient that empirically set as 0.1. Combining SSP and PMP, our pre-training loss is a multi-task learning loss with the mixed training loss $L_{all}=L_{S\!S\!P}+L_{P\!M\!P}$.

By combining two pre-training strategies SSP and PMP, the solver strengthens the cognition of complex geometry layout. In SSP, the modeling of local relationships leads to the global relationship understanding, for example, we can infer that the mask token in the semantic clause ``BD $\perp$ EA on [M]” is ``C” according to structural clauses ``line B C D” and ``line E C A”. Via PMP, the textual points become aware of layout position from positional encoded image patches by alignment. We do not adopt the simple and direct way of fine-grained 2D position embedding such as in LayoutLM \cite{Xu2020, Xu2021}. This is because existing GPS datasets do not support large-scale layout understanding pre-training. It is also akin to human geometric cognition in that accurate positioning is not required to understand geometry layout.

\subsection{Layout-Aware Fusion Attention}
Although LANS has already acquired a certain level of layout understanding through the pre-training strategies above, this ability can fade to some extent during downstream training because of the different training targets of GPS. To compensate for the loss of layout awareness in the GPS training phase, we propose layout-aware fusion attention (LA-FA) to enhance the intra-modal and cross-modal token fusion. LA-FA is located between the bidirectional GRU encoder and the self-limited GRU decoder. 

\begin{figure}[t]
    \begin{center}
    \includegraphics[width=1.0\columnwidth]{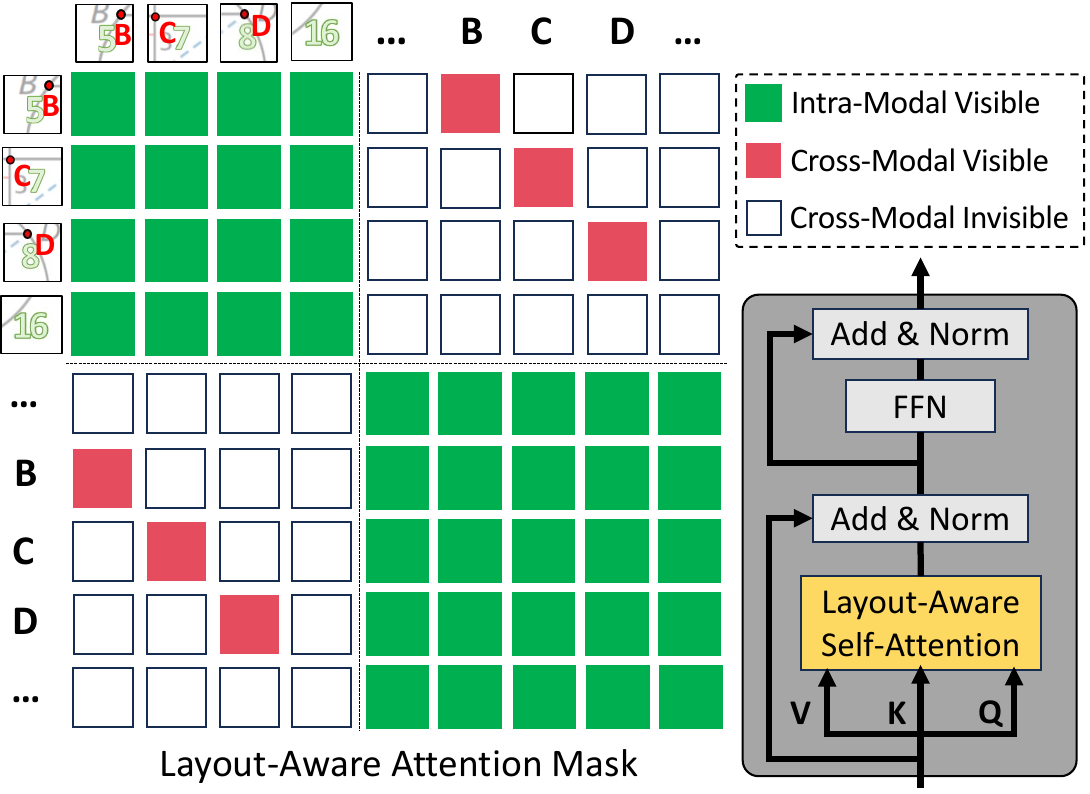} 
    \end{center}
    \vspace{-0.2cm}
    \caption{Schematic of Layout-Aware Fusion Attention.}
    \vspace{-0.2cm}
    \label{fig:attention}
\end{figure}

As shown in Figure \ref{fig:attention}, the LA-FA module is similar to the transformer encoder block \cite{Vaswani2017} which also contains layer normalization, feed-forward layer,  and residual connection except the layout-aware self-attention. Our layout-aware self-attention uses the carefully designed layout-aware attention mask which allows visibility to all intra-modality tokens but restricts cross-modality visibility if the textual point is not inside the image patch in the visual space. Specifically, we construct the mask matrix $M_{i,j}\;(1 \!\leq\! i,j\!\leq\! N_D\!+\!N_T)$, which consists of value 0 as invisible and value 1 as visible:
\begin{equation}
\textstyle
M_{i,j} \!=\! 
\left\{
\begin{array}{ll}
    1, & \text{if } (i, j) \in V\!V  \\
    1, & \text{if } (i, j) \in T\!T \\
    1, & \text{if } (i, j) \in V\!T \& \text{Pos}(t_j) \in \text{Reg}(d_i) \\
    0, & \text{otherwise}
\end{array}
\right.
\end{equation}
where $V\!V = \{(i, j) \,|\, 1 \! \leq \! i, j \!\leq\! N_D \}$ is the mask region of visual intra-modality, $T\!T = \{(i, j) \,|\, N_D\!\!+\!\!1\!\leq\! i, j \!\leq\! N_D\!\!+\!\!N_T\}$ is the mask region of textual intra-modality, $V\!T = \{(i, j) \,|\, 1 \!\leq\! i, j \!\leq\! N_D\!+\!N_T\}\!-\!V\!V\!-\!T\!T$ is the mask region of cross-modality, $\text{Pos}(t_j)$ denotes the visual position of point token $t_j$ and $\text{Reg}(d_i)$ refers to the visual region of image patch $d_i$. Moreover, layout-aware fusion-attention (LA-FA) could be computed by:
\begin{equation}
\text{LA-FA}(Q,K,V,M) \!=\! \text{softmax}\!\left(\frac{QK^T}{\sqrt{m_k}}\cdot M\!\right)V
\end{equation}
where $Q,K,V$ are query matrix, key matrix, and value matrix all transformed from encoding context $H$, and $m_k$ is the dimension of the key vector.

In summary, in the process of cross-modal fusion, LA-FA leverages the point position to guide the attention between diagram and text, strengthening the understanding of diagram layout. For mitigating the optimization burden, we only use one LA-FA block, as adding more blocks does not bring extra improvement according to our experiments.

\begin{table*}[t]
\centering
    \resizebox{0.95\linewidth}{!}{
    \begin{tabular}{lccccccc}
        \toprule
        \multirow{2}{*}{Method}  & \multicolumn{3}{c}{Geometry3K} & \multicolumn{3}{c}{PGPS9K} & \multirow{2}{*}{Parameters} \\
        \cmidrule(lr){2-4} \cmidrule(lr){5-7}
                    & Completion & Choice & Top-3 & Completion & Choice & Top-3  \\
        \midrule
        Human Expert \cite{Lu2021}   & - & \textit{90.9} & - & - & - & - & - \\
        InterGPS (Predict)* \cite{Lu2021} & 44.6  & 56.9 & - & - & - & - & -  \\
        InterGPS (Diagram GT)* \cite{Lu2021} & 64.2  & 71.7 & - & 59.8 & 68.0 & - & -  \\
        InterGPS (All GT)* \cite{Lu2021} & 69.0 & 75.9 & - & -  & -  & - & - \\
        GeoDRL (Predict) \cite{Peng2023} & - & 68.4 & - & -  & -  & - & - \\
        \midrule
        Baseline (Neural Solver) \cite{Lu2021} & - & 35.9 & -  & - & - & - & -  \\
        NGS$^\&$ \cite{Chen2021} & 35.3 & 58.8 & 62.0 & 34.1 & 46.1 & 60.9 & 80M \\
        Geoformer$^\&$ \cite{Chen2022} & 36.8 & 59.3 & 62.5 & 35.6 & 47.3 & 62.3 & 267M \\
        SCA-GPS \cite{Ning2023} & - & 76.7 & -  & - & - & -  & > 310M \\
        PGPSNet \cite{Zhang2023} & 65.0 & 77.9 & 80.7  & 62.7 & 70.4 & 79.5  & 23M  \\
        \midrule
        LLaVA-v1.5 \cite{liu2023visual} & 7.6 & 11.2 & -  & 6.3 & 9.1 & - & 7B    \\
        mPLUG-Owl2 \cite{ye2023mplugowl2} & 12.1 & 17.4 & -  & 10.1 & 13.1 & - & 7B  \\
        Qwen-VL \cite{bai2023qwen} & 22.1 & 26.7 & -  & 20.1 & 23.2 & -  &  7B \\
        GPT-4V \cite{openai2023gpt4} & 38.7 & 41.4 & -  & 30.2 & 35.7 & -  & - \\
        \midrule
        LANS (ours)  & \textbf{72.1} & \textbf{82.3} & \textbf{82.8}  & \textbf{66.7} & \textbf{74.0} & \textbf{82.2} & 26M \\
        \bottomrule
    \end{tabular}
    }
\caption{Performance comparison among state-of-the-art GPS solvers. * denotes results re-produced with the open source code. $\&$ denotes methods re-implemented by us.
}
\label{tab:performance_compare}
\end{table*}

\section{Experiments}
\subsection{Setup}
\paragraph{Model Architecture} 
The patch projection module for diagram chooses the CNN architecture, selecting a light-weight ResNet10 \cite{he2016deep} to extract feature map before meshing. Feeding with diagram images resized as 256$\times$256, the patch projection module maps diagram into 8$\times$8=64 image patches. In default, we employ a 6-layer, 8-head, 256-input, and 1024-hidden dimensional transformer \cite{Vaswani2017} as the architecture of MLA-PLA, and a multi-head attention with the same head number and feature dimension for LA-FA. The bidirectional GRU encoder and self-limited GRU decoder in LANS are adopted following the same architecture as PGPSNet \cite{Zhang2023}. Besides, a dropout layer with the value 0.2 is added behind the patch projection module to prevent overfitting during the training stage. 
\paragraph{Training Hyperparameters Details} 
\label{sec:eval metrics}
We choose the AdamW optimizer \cite{Loshchilov2017DecoupledWD} with the weight decay $1\times10^{-2}$ and the step decline schedule with the decay rate of 0.5, and the training batch size is set as 128. We provide a more detailed description of the remaining parameters we use during the pre-training and fine-tuning stages in the appendix \ref{sec:appendix_train_detail}. 
\paragraph{Datasets and Metrics} 
We evaluate the performance of proposed LANS on two plane geometry problem datasets: Geometry3K \cite{Lu2021} and PGPS9K \cite{Zhang2023}. They all have fine-grained diagram annotation and interpretable solution programs. The textual clauses and point positions used in this paper are converted from the diagram annotation. The solution program consists of several solving steps, each step consists of an operator and associated operands, where the operator corresponds to a geometric theorem and operands are arranged according to the theorem formula. The paired program executor based on Python calculates the numerical results of solution programs. The MLA-PLA module of LANS is pre-trained from scratch on PGPS9K dataset that masks solution programs, because of the shortage of geometric corpus and the great distribution gap in contrast with natural corpus. 

Similar to PGPSNet \cite{Zhang2023}, we use three evaluation metrics to assess the numerical performance of our LANS, namely \textit{Completion}, \textit{Choice}, and \textit{Top-3}. In the Completion, the neural solver selects the first executable solution program as the Completion result. The Choice is defined as choosing the correct option from four candidates but selecting one randomly if the outputted answer is not in. In the Top-3, the solution is considered correct if it is among the top three confidence solutions. We set the \textit{Completion} as evaluation metric for ablation study in section \ref{ablation} by default.  Given the outstanding capabilities of multimodal large models in addressing multimodal reasoning problems, we compared popular existing open-source multimodal large models in Table \ref{tab:performance_compare} with the currently most powerful multimodal model, GPT-4V. Evaluation was conducted in both Completion and Choice modes, where in Completion mode, the large model was required to directly provide answers, and in Choice mode, reference options were added to the prompt for the large model.

\subsection{Comparison with State-of-the-art Solvers}
\begin{figure*}[t]
    \begin{center}
    \includegraphics[width=1.95\columnwidth]{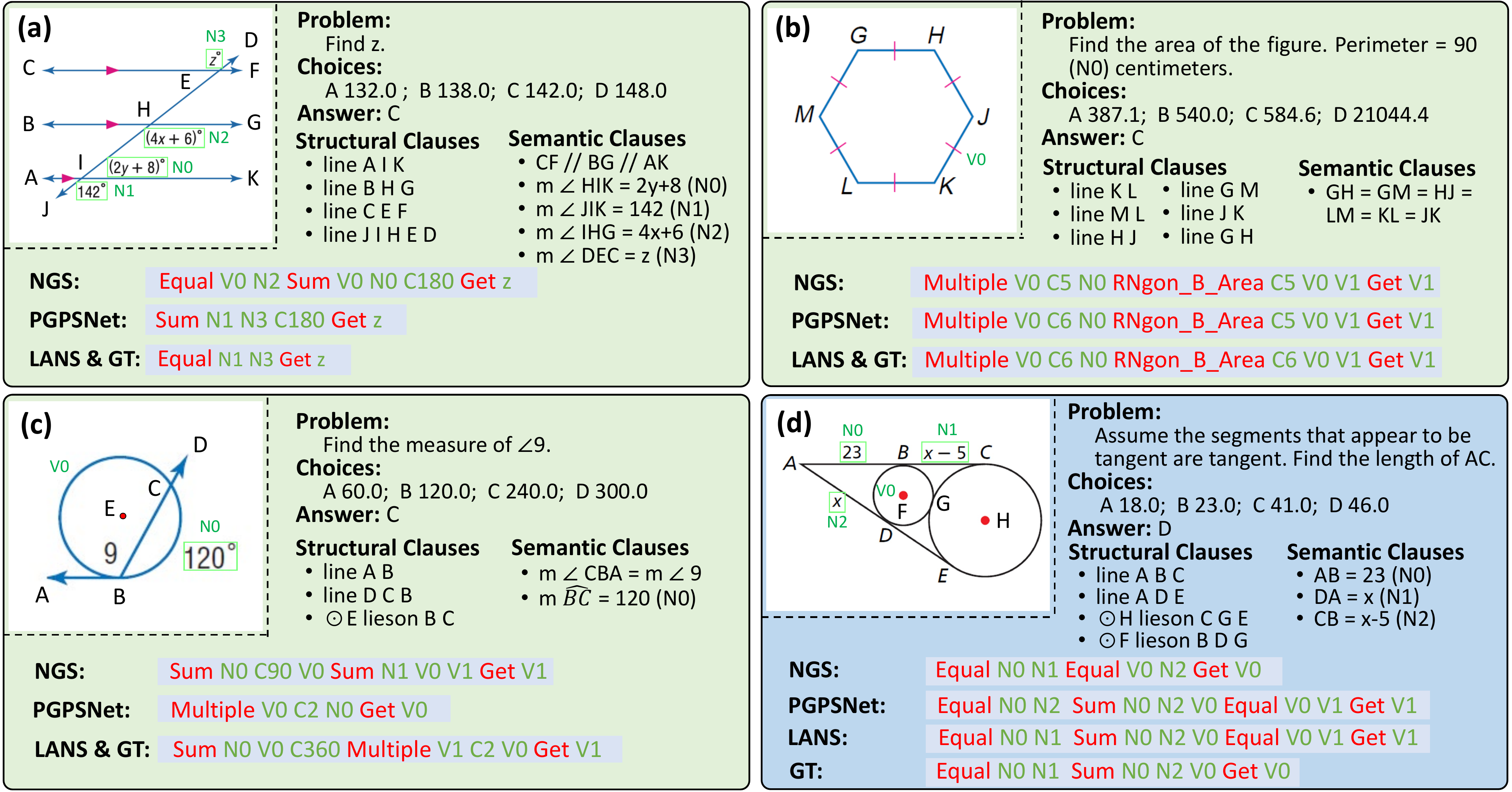} 
    \end{center}
    \vspace{-0.2cm}
    \caption{Case analysis on PGPS9K. Solving above problems requires layout awareness of geometry diagram. (a), (b) and (c) are the problems LANS answered correctly, (d) is the problem LANS answered incorrectly.}
    \vspace{-0.2cm}
    \label{fig:case}
\end{figure*}
We compare LANS with state-of-the-art models, including neural solvers, symbolic solvers, and multimodal large models in Table \ref{tab:performance_compare}, in terms of both performance and parameter quantity. The results indicate that our LANS achieves excellent model performance by incorporating efficient parameters.

As to symbolic solvers, InterGPS \cite{Lu2021} solved geometry problems by searching and matching with unified formal language. According to the input source of formal language, InterGPS presents three types of results, e.g., ``Predict" means that all formal language is predicted by its parsers, ``Digram GT" denotes that formal clauses of diagram use ground truth, and ``All GT" indicates that formal clauses of diagram and textual problem are all ground truth. GeoDRL \cite{Peng2023} improved the search strategy of Inter-GPS with logical graph deduction and deep reinforcement learning. Experimental results show that our LANS outperforms symbolic solvers on all datasets and in all evaluation metrics. Even compared with InterGPS (All GT) which uses annotated formal clauses designed carefully, LANS gains a 3.1\% improvement in Completion and a 6.4\% improvement in Choice mode on Geometry3K Dataset.   

As to neural solvers, NGS \cite{Chen2021} and Geoformer \cite{Chen2022} relied primarily on textual problems to solve problems. Even though re-implementing them with the textual clauses parsed from the diagram and the same augmentation strategies, performance gaps between these two solvers and our LANS are still significant, 32.6\% and 31.1\% lower in Completion on PGPS9K, respectively. SCA-GPS \cite{Ning2023} shows similar performance as InterGPS (All GT) because diagram understanding methods, character alignments, and masked image modeling, are coarse-grained and ineffective. PGPSNet \cite{Zhang2023} employed textual clauses to model diagram layout but lost lots of visual information. Our LANS is enhanced at modal alignment and fusion for better layout awareness and surpasses PGPSNet by 7.1\% and 4.0\% in Completion on Geometry3K and PGPS9K. The improvements in Top-3 are less than in Completion because most of the correct solutions are concentrated among highly confident candidates.  

Our approach far surpasses the performance of current multimodal large models. This may be attributed to the presence of complex symblic OCR information, layout details, and abstract elements in geometric images, where the perception capabilities of the Gemini \cite{team2023gemini} and GPT-4V \cite{achiam2023gpt} models are insufficient. Similar phenomena have also been observed in the MathVista \cite{lu2023mathvista} benchmark. For evaluation results and detailed information on the Completion and Choice multimodal large-scale models, please refer to section \ref{sec:appendix_eval_detail}.
 
\subsection{Ablation Study} \label{ablation}
\paragraph{Effect of Modules} 
To examine the effect of our proposed modules in LANS, we conducted ablation experiments on the Geometry3K dataset, taking PGPSNet solver \cite{Zhang2023} who owns the SS-PLM module but without the LA-FA module as the baseline. Experimental results presented in Table \ref{tab:module} illustrate that MLA-PLM module with multimodal pre-training is superior to SS-PLM module with only text-modal pre-training and obtains a 5.4\% improvement. LA-FA module further boosts GPS via multi-modal feature fusion in the training phase and achieves a 72.1\% accuracy, over baseline 7.1\%.
\begin{table}[H]
    \centering
    \begin{small}
    \begin{tabular}{p{3.5cm}c}
        \toprule
        Module  & Accuracy  \\
        \midrule
        Baseline      &  65.0            \\
        + MLA-PLM     &  70.4 \footnotesize{(\textcolor{red}{+5.4})}     \\
        + MLA-PLM + LA-FA  &   \textbf{72.1} \footnotesize{(\textcolor{red}{+7.1})}   \\
        \bottomrule
    \end{tabular}
    \end{small}
    \caption{Ablation study of modules on Geometry3K.}
    \label{tab:module}
\end{table}
\paragraph{Role of Pre-training Strategies} To validate the role of pre-training strategies within MLA-PLM, we did ablation experiments on both SSP and PMP pre-training strategies. Ablation experiments involved two processes: first pre-training with various strategies and then fine-tuning on Geometry3K. Table \ref{tab:ablation_pretrain} verifies that SSP and PMP pre-training strategies all improve GPS, where SSP promotes global relationship recognition and PMP aligns visual points and textual points. The comparison between row 2, row 3, and row 4 demonstrates that the combination of SSP and PMP realizes complex layout understanding synthetically, thus promoting problem-solving together. 
\begin{table}[H]
    \centering
    \begin{small}
    \begin{tabular}{p{3.5cm}c}
        \toprule
        Pre-training Strategy  & Accuracy \\
        \midrule
        None & 38.2   \\
        + SSP  &  55.4 \footnotesize{(\textcolor{red}{+17.2})}          \\
        + PMP  &  66.9 \footnotesize{(\textcolor{red}{+28.7})}          \\
        + SSP + PMP    &   \textbf{72.1} \footnotesize{(\textcolor{red}{+33.9})}        \\
        \bottomrule
    \end{tabular}
    \vspace{-0.2cm}
    \end{small}
    \caption{Ablation study of pre-training strategies on Geometry3K dataset.}
    \label{tab:ablation_pretrain}
\end{table}
\paragraph{Role of Attention Mask} To validate the role of attention mask within the LA-FA module, we compare three types of attention masks: w/o LA-FA, vanilla attention mask  \cite{Vaswani2017}, and layout-aware attention mask. Compared with the vanilla attention mask with global visibility, layout-aware attention mask guided by point positions promotes modal fusion and strengthens diagram understanding. The results in Table \ref{tab: attention} also indicate the significance of layout-aware attention.
\begin{table}[H]
    \centering
    \small
    \begin{tabular}{lc}
        \toprule
        Mask Type  & Accuracy  \\
        \midrule
        w/o LA-FA  &  70.4    \\
        w Vanilla Attention Mask &  70.6    \\
        w Layout-Aware Attention Mask &  \textbf{72.1}  \\
        \bottomrule
    \end{tabular}
    \caption{Ablation study of attention mask on Geometry3K dataset.}
    \label{tab: attention}
\end{table}

\subsection{Case Analysis and Fail cases}
We also conducted a case analysis to discuss the strengths and weaknesses of solvers. Figure \ref{fig:case} displays four plane geometry problems (a)-(d) involving various geometric layouts, and they rely on good layout awareness to solve them. In case (a), the position of C relative to F determines if $\angle$JIK and $\angle$DEC are corresponding or alternate angles. Results show LANS identifies corresponding angles accurately, unlike other solvers. In case (b), the perception of polygon edge number is the key to solving this problem. Contrary to LANS, other solvers cannot count edge numbers correctly through the diagram or textual clauses, resulting in a wrong solution. Case (c) is the same problem as shown in Figure \ref{fig:example} in which textual clauses cannot identify diagram uniquely. In contrast with PGPSNet, LANS can judge the orientation and type of $\angle$ABC and get the right solution.  

\section{Conclusion}
We propose a layout-aware neural solver LANS to understand complex layouts of plane geometry diagrams. Benefiting from the multimodal layout-aware pre-training, LANS is endowed with abilities of global relationship cognition and cross-modal point alignment. Thanks to layout-aware fusion attention, LANS further improves cross-modal fusion directed by point positions. The experimental results demonstrate the superiority of LANS enhanced with layout awareness. 
\section*{Limitations}
LANS is still limited to point primitives to carry out layout understanding. In the future, we will try to align higher-level geometric primitives to obtain better layout understanding and modal fusion. Besides, LANS may generate redundant solution sequences. Case (d) in Figure \ref{fig:case} is a complex layout scenario that none of the solvers can solve correctly. In conclusion, the case analyses above fully indicate that LANS promotes GPS with enhanced layout awareness. Integrating richer layout information and symbolic cues of elements through multimodal pretraining is a direction worthy of further exploration.

\section*{Ethical Impact}
As a neural solver addressing multimodal mathematical problems, LANS has the potential for application in educational settings, specifically for the automatic resolution of mathematical problems. This utilization can contribute to promoting educational equity.

\bibliography{acl_latex}
\appendix
\label{sec:appendix}
\section{Dataset Details}
We evaluated our method on two datasets, Geometry3K and PGPS9K, each containing high-quality diagram images. Table \ref{tab:data_detail_appendix} provides a comparison of key information between the two datasets, with PGPS9K featuring a wider variety of question types and higher reasoning complexity.
\begin{table}[h]
\small
\centering
\begin{tabular}{lccccc}
\toprule
Dataset       & \#QA & \#Type &\#Avg OP &\#Avg PL \\
\midrule
Geometry3K    & 3,002  & 4 & 1.98 & 5.35\\
PGPS9K        & 9,022  & 30 & 2.43 & 7.45\\
\bottomrule
\end{tabular}
\caption{Type, OP and PL represent problem type, operator number and program length, respectively}
\label{tab:data_detail_appendix}
\end{table}

\section{Training Details}
To ensure the reproducibility of the paper, we provide here the key hyperparameters used during training, as well as the data augmentation methods employed. Additionally, within our method framework, how Patch Projection is relied upon and the granularity of Patch division are crucial for achieving the effectiveness of our approach as described in the paper. We discuss here the impact of these parameters on the replicability of the model.
\subsection{Optimzation Parameters Details}
\label{sec:appendix_train_detail}
During the pre-training phase, the learning rate is initialized to $5\times10^{-4}$ and the learning rate decay is applied at 1,000, 1,800, 2,400, and 3,000 epochs with a total of 3,500 epochs. During the training stage, all modules of LANS train together with an initial learning rate as $1e^{-4}$ for language model MLA-PLM and $1e^{-3}$ for other modules, decaying at 160, 280, 360, 440 and 500 epochs uniformly with a total 520 epochs.  

All experiments were conducted on an 8-GPU Titan XP server. Training of the MLA-PLM module took approximately 20 hours on a 4-GPU machine, while fine-tuning of LANS on 4 GPUs took 8 hours.
\subsection{Data Augmentation Details}
We scale the image to 256 on the longest side and place it in the center of 256$\times$256 blank screen. The diagram is flipped randomly and changes the point positions accordingly. For text, following the work \cite{Zhang2023}, we apply four augmentation strategies: token replacement, connection rotation, representation transposition, and clauses shuffle. These augmentation strategies not only improve the diversity of geometry problems but also provide geometric solvers with basic geometric representation knowledge. 
\subsection{Impact of hyperparameters}
\noindent \textbf{Discussion on the Granularity of Patch Division}.
To assess the influence of image patches, we adopted four configurations of patch numbers: 1$\times$1, 4$\times$4, 8$\times$8, and 16$\times$16. In Table \ref{tab:patches}, we observe that LANS benefits from fine-grained partitions of the diagram, based on the comparison of row 1 with rows 2, 3, and 4. However, according to the comparison of row 3 with row 4, problem-solving performance declines if the diagram is over-segmented. The possible explanation is that redundant and blank image grids, which are generated from patch partition, interfere with model attention while increasing the burden of model computation. Therefore, considering overall performance and speed, we choose the 8$\times$8 configuration as our model setup. \\
\begin{table}[H]
    \centering
    \begin{small}
    \begin{tabular}{lcc}
        \toprule
        Image Patch Num.  & Geometry3K & PGPS9K \\
        \midrule
        1 $\times$ 1      &  65.0   &  62.7  \\
        4 $\times$ 4      &  70.5    &   \textbf{66.8}    \\
        8 $\times$ 8     &  \textbf{72.1}  & 66.7 \\
        16 $\times$ 16    &  69.1  & 65.4 \\
        \bottomrule
    \end{tabular}
    \end{small}
    \caption{Comparison of Different Image Patch Numbers.}
    \label{tab:patches}
\end{table}

\noindent \textbf{Discussion on the Projection Method of Image Patches}. To validate the impact of patch projection schemes, in Table \ref{tab:projection}, we tested three types of patch projection modules: None, linear layer, and CNN model. None refers to not using the patch projection module and also not inputting image patches. In our experiments, we find that a redundant placeholder in None does harm to GPS due to additional meaningless optimizations. The linear-based patch projection maps image grids linearly and produces corresponding image patches, which is also commonly adopted in recent transformer architectures \cite{Kim2021,pmlr-v162-li22n}. However, this module does not fit to geometry diagram because it may damage the geometric structure. CNN-based patch projection first extracts global features and then mesh feature maps. That module could better understand the overall layout, bringing with higher solving performance and more stable training, and it is also set as the default patch projection module.

\begin{table}[H]
    \centering
    \begin{small}
    \begin{tabular}{lcc}
        \toprule
        Projection Type  & Geometry3K & PGPS9K \\
        \midrule
        None &  64.2  &   61.3      \\
        Linear   &  69.4  &  65.5   \\
        CNN   &  \textbf{72.1}  &   \textbf{66.7}     \\
        \bottomrule
    \end{tabular}
    \end{small}
    \caption{Comparison of different patch projections.}
    \label{tab:projection}
\end{table}

\begin{table*}[th!]
\centering
\small
\renewcommand\tabcolsep{2.9pt} 
\begin{tabular}{c|p{12cm}}
 \toprule
 \textbf{Eval Mode} & \textbf{Prompt}\\
 \midrule
\textit{Choice}  & 
 \begin{minipage}[s][2.5cm]{1.5\columnwidth}
 \textbf{Role Prompt:} You are a geometric problem-solving robot. Please solve the following geometry problems based on the contents of the diagram and the problem description. \\
\textbf{Diagram}:  The Diagram is <img>{images/img\_3755.png}</img>{} \\
 \textbf{Question:} If RL = 5, RT = 9, and WS = 6, find RW.\\
\textbf{Choices:} (A) 5.4 (B) 6.6 (C) 6.0 (D) 7.5 \\
 \textbf{Format Prompt:} Please give reason process and provide the correct option, such as: the answer is A/B/C/D:.
\\
 \end{minipage}
 \\
 \midrule
\textit{Completion}  & 
 \begin{minipage}[s][2.5cm]{1.5\columnwidth}
 \textbf{Role Prompt:} You are a geometric problem-solving robot. Please solve the following geometry problems based on the contents of the diagram and the problem description. \\
 \textbf{Diagram}:  The Diagram is <img>{images/img\_2056.png}</img>{} \\
 \textbf{Question:} 
 Polygon $ABCD \sim AEFG$, $ \angle AGF = 108^\circ$, $GF = 14$, $AD = 12$, $DG = 4.5$, $EF = 8$, and $AB = 26$. Find $ \angle ADC$.\\
\textbf{Format Prompt:} Please give reason process and provide the correct option, such as: the answer is 15.0:.
\\
\end{minipage}
\\
\bottomrule
\end{tabular} 
\captionsetup{justification=centering}
\caption{The prompts used for Choice and Completion Modes in two specific questions.}
\label{tab:prompt_for_mllm}
\end{table*}

\begin{table}[th!]
\small
\centering
\begin{tabular}{p{0.17\linewidth}p{0.3\linewidth} p{0.4\linewidth}}
\toprule
\textbf{Model Name} & \textbf{Model Repository Name/API Version} & \textbf{Sampling Parameters} \\
\midrule
Qwen-VL & Qwen/Qwen-VL-Chat & do\_sample = True, top-k = 5, max\_length = 512 \\
\midrule
LLaVA-1.5 & liuhaotian/llava-v1.5-13b & do sample = True, temperature = 0.2, max new tokens = 1024\\
\midrule
mPLUG-Owl2 & MAGAer13/mplug-owl2-llama2-7b & do sample = True, top-k = 5, max length = 512\\ 
\midrule
GPT4V & gpt-4-1106-vision-preview & Chatbot URL: \url{https://chat.openai.com} \\ 
\bottomrule
\end{tabular}
\caption{Generating parameters and Huggingface model repository names for multimodal large models}
\label{tab:lmm_generating_params}
\end{table}

\begin{figure*}[t]
    \begin{center}
    \includegraphics[width=2.0\columnwidth]{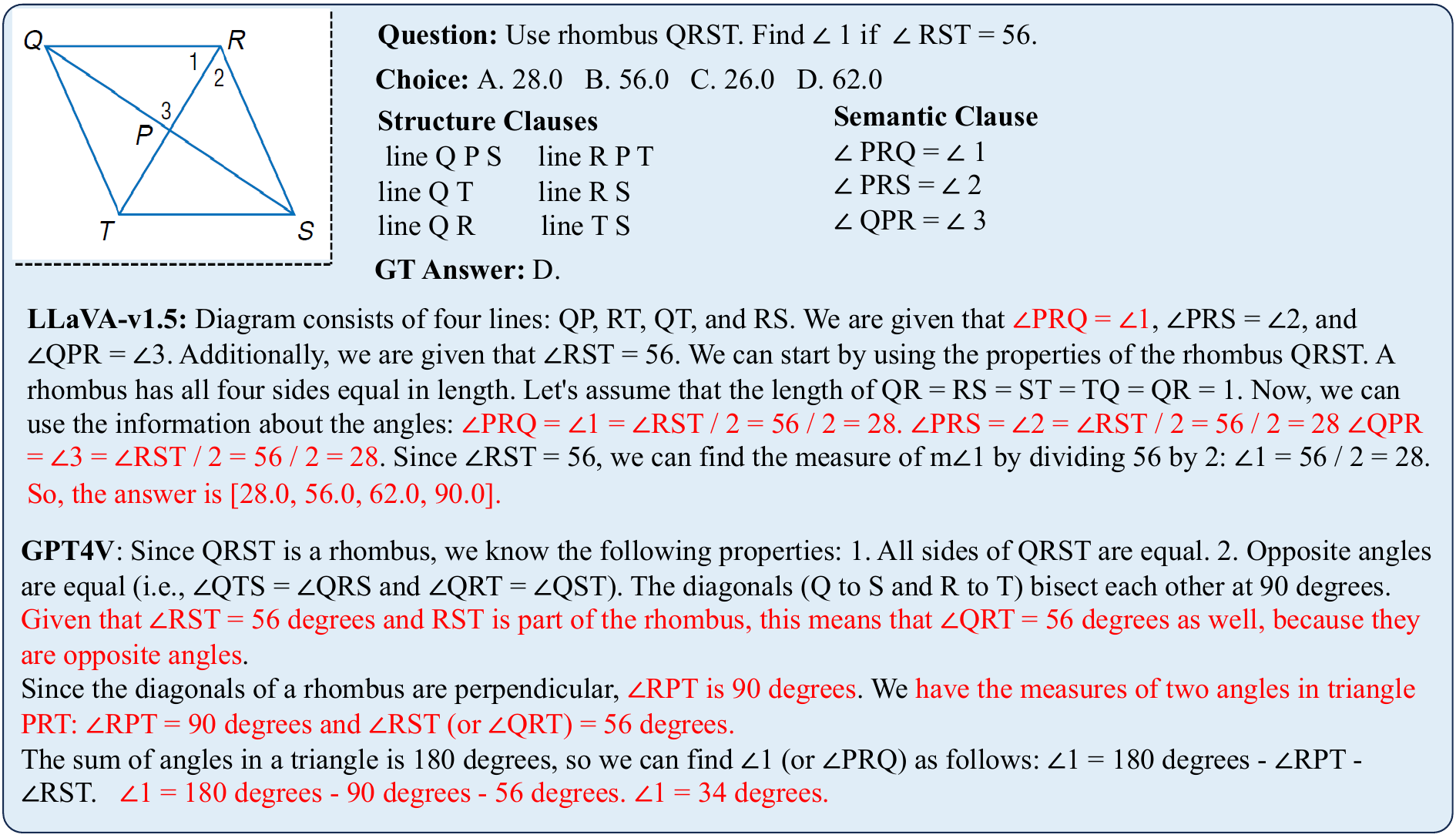} 
    \end{center}
    \caption{Case of the multimodal large model LLaVA-v1.5 and GPT4V. The red ones are marked as generated inference hallucinations \cite{Zhang2023}.}
    \label{fig:mllm_case}
\end{figure*}

\section{Mulitmodal LLM Eval Details}
\label{sec:appendix_eval_detail}
\subsection{Eval Prompt Details}

We illustrate in Table \ref{tab:prompt_for_mllm} with examples of how the prompts used for evaluating the multimodal large model vary across different Eval Mode. Our Prompt consists of several components, including \textit{Role Prompt}, \textit{Diagram}, \textit{Question}, \textit{Choices}, and \textit{Format Prompt}. The \textit{Role Prompt} specifies the type of problem the large model is tasked with solving and the actions it needs to perform. \textit{Diagram} depicting the form's content, textual description of the \textit{Question}, and \textit{Choices}. To ensure the large model generates standardized output for easy scoring, we have set a \textit{Format Prompt}. The main difference in evaluation between \textit{Choice} and \textit{Completion} modes lies in that, in \textit{completion} mode, the prompt does not provide reference options.

\subsection{Generation Parameter Details}
We list the relevant important parameters used for evaluation across different multimodal large models in Table \ref{tab:lmm_generating_params}.

\subsection{Output detail and  discussion of MLLM}
Table \ref{tab:performance_compare} demonstrates that multimodal large models perform poorly in solving geometric problems. Some open-source models exhibit performance lower than random guessing for select questions. We attribute this to the fact that the visual comprehension component of current multimodal large models is primarily tailored to natural scene images and struggles with abstract forms. For instance, in Figure \ref{fig:mllm_case}, both GPT4V and LLaVA-v1.5 fail to grasp the relational elements within geometric diagrams, such as understanding the reference to Angle 1. Additionally, models like LLaVA-v1.5 may also generate severe model hallucinations, even output in an incorrect format like "So, the answer is [28.0, 56.0, 62.0, 90.0]," which prevent the extraction of correct results and result in low accuracy.
\end{document}